\documentclass[10pt,twocolumn,letterpaper]{article}

\usepackage{cvpr} 
\usepackage{times}
\usepackage{graphicx}
\usepackage{amsmath}
\usepackage{amssymb}
\usepackage{booktabs}
\usepackage{multirow}
\usepackage{xcolor}
\usepackage{hyperref}
\usepackage{cleveref}
\usepackage{pifont}
\usepackage[acronym]{glossaries}
\newacronym{FRS}{FRS}{Face Recognition System}
\newacronym{XAI}{XAI}{Explainable AI}
\newacronym{VLM}{VLM}{Vision-Language Model}
\newacronym{LLM}{LLM}{Large Language Model}
\newacronym{EER}{EER}{Equal Error Rate}
\newacronym{FTA}{FTA}{Failure to Acquire}
\newacronym{FISWG}{FISWG}{Face Identification Scientific Working Group}
\newacronym{ENFSI}{ENFSI}{European Network of Forensic Science Institutes}

\usepackage{graphicx}
\usepackage{subcaption}
\usepackage{comment}

\usepackage[table]{xcolor}

\usepackage{booktabs}
\usepackage{tabularx}
\usepackage{listings}
\usepackage{xcolor}

\begin{document}

\title{Benchmarking the Explanatory Quality of Open-Weight Vision-Language Models in Face Recognition
}

\author{\large Laurent Colbois\textsuperscript{1} and Sébastien Marcel\textsuperscript{1,2}\\%
\textsuperscript{1} \small{Idiap Research Institute, Switzerland}\\%
\textsuperscript{2} \small{Université de Lausanne, Switzerland}\\%
{\tt\small \{laurent.colbois, sebastien.marcel\}@idiap.ch}
}

\maketitle

\thispagestyle{empty}
\vspace{-10cm}

\begin{abstract}
\label{sec:abstract}

    \Glspl{VLM} have recently been proposed as promising tools for face recognition, as they can produce natural language explanations alongside similarity scores. This capability is considered appealing for face comparisons in forensic contexts, which require decisions to be transparent and auditable. However, existing evaluations of \glspl{VLM} for that use case focus mostly on recognition accuracy, while the validity of generated explanations remains unquantified.
In this work, we introduce a benchmarking framework for VLM-based face recognition that treats explanation quality as a core evaluation axis. We propose two criteria that explanations should satisfy: relevance, i.e., reliance on identity-stable facial features; and faithfulness, i.e., alignment with the visible image content without hallucinated features. We jointly develop a methodology enabling the quantification of relevance and faithfulness of evaluated models, based on constraining model outputs to a structured explanation format that supports automated querying and auditing.
Using this framework, we benchmark several families of open-weight VLMs, jointly evaluating face verification accuracy and explanation quality. Our results highlight remaining shortcomings of produced explanations, and emphasize the need for such explanation quality metrics to get a complete picture of model performance. The proposed benchmark and open-source evaluation harness provide a foundation for proper benchmarking and future fine-tuning of explainable face recognition systems.

\end{abstract}

\section{Introduction}

\begin{figure}[htbp]
\centering
\includegraphics[width=0.8\columnwidth]{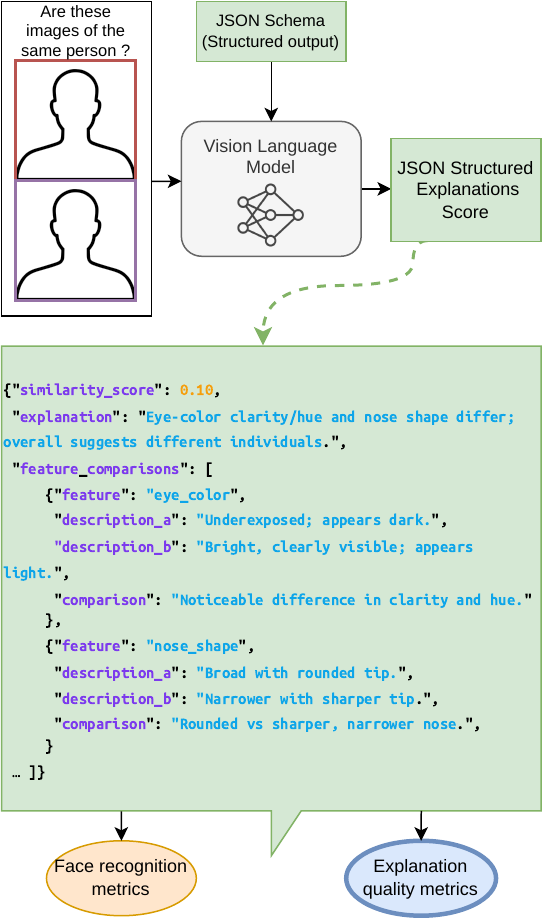}
\caption{We prompt \glspl{VLM} with pairs of face images and a question requiring them to assess whether these images represent the same person or not. Moreover, the model is forced to structure its answer following a predefined JSON schema, thus enabling facilitated queryability of the generated explanations for automated scoring.}
\label{fig:project schematic}
\end{figure}

Automated \glspl{FRS} have evolved into reliable tools for verification tasks in a variety of real-world applications, such as border control or mobile device security. They have been demonstrated to provide high levels of accuracy in the vast majority of situations, delivering convenience while maintaining a low false accept rate. Nevertheless, this high level of statistical performance does not yet meet the standard required for high-stakes, case-specific decisions, as is often the case in forensic face comparison tasks. In this context, the black-box nature of the neural networks used as underlying recognition models limits their explainability, challenging the applicability of \glspl{FRS} by preventing the creation of auditable reports required by forensic standards as part of biometric analyses.

\Gls{XAI} has been a longstanding research field, including works that aim to increase the explainability of \glspl{FRS}. Early approaches have predominantly focused on the development of \emph{a posteriori} pixel-based saliency maps~\cite{selvarajuGradCAMVisualExplanations2017}, which provide insights into the influence of specific regions of the input face images on the final face similarity score. More recently, a novel line of work has emerged, enabled by the increasing availability of multimodal \glspl{VLM}. Early studies have demonstrated the solid zero-shot capabilities of \glspl{VLM} for face recognition~\cite{sonyBenchmarkingFoundationModels2025,otroshishahrezaFoundationModelsBiometrics2025}. While their recognition accuracy has not yet reached the state of the art achieved by dedicated \glspl{FRS}, it remains relatively high even in zero-shot approaches. More importantly, \glspl{VLM} enable the face similarity score to be accompanied by text-based explanations, which are much closer to human-written reports than saliency maps and may therefore appear less abstract to the forensic community.

However, we argue that the mere presence of text-based explanations is not, by itself, a guarantee of proper explainability. For forensic use, explanations must satisfy constraints that make them actually valid and valuable. In particular, explanations should avoid reliance on transient evidence (e.g., glasses, facial expression, lighting), and should not introduce unsupported claims by describing facial attributes that are not visible in the images (e.g., due to occlusions). Unfortunately, prior works generally do not discuss the validity of produced explanations beyond brief qualitative observations. In this work, we propose that explanation quality should be treated as a key evaluation axis in the benchmarking of \glspl{VLM} for forensic face recognition, alongside recognition accuracy.

Our aim is therefore to develop a benchmarking methodology that jointly evaluates recognition accuracy and explanation quality for \gls{VLM}-based face verification, and to apply it to a range of open-weight models. Central to our approach is to constrain model outputs to a structured format that enables automated querying and auditing at scale. While several properties contribute to explanation quality (including factual correctness with respect to attribute annotations), this paper focuses on two dimensions that can be evaluated in a dataset-driven and auditable manner: \textbf{relevance} (exclusive reliance on identity-stable evidence) and \textbf{faithfulness} (avoidance of unsupported claims about non-visible content).
Our main contributions are the following:
\begin{itemize}
\item We introduce two forensic-motivated explanation criteria for VLM-based face recognition: \textbf{relevance}  and \textbf{faithfulness}.
\item We propose a method to quantify these criteria by constraining outputs to a structured format that supports automated querying and auditing.
\item We benchmark multiple open-weight VLM families across scales, evaluating face verification performance alongside relevance and faithfulness, and analyze the effects of constrained decoding and model scaling.
\end{itemize}

The code required to reproduce our experiments, including our benchmarking framework, is released publicly\footnote{\url{https://gitlab.idiap.ch/biometric/vlmfr}}. We hope it can serve as an evaluation harness for future works focusing on improving \gls{VLM}-based face recognition while preserving explanation quality.

\section{Related Work}

\subsection{Guidelines for forensic face comparisons}
The practice of face comparison by human experts in forensic contexts predates the era of automated face recognition and remains prevalent today.
Given the sensitivity of such analyses, the forensic community has devoted substantial effort to establishing standards that support consistent and rigorous practice.
In particular, the \gls{FISWG} has issued standards for morphological analysis of human faces \cite{FacialImageComparison2018} and for assessing the stability of facial features among adults \cite{PhysicalStabilityFacial2021}.
These standards are recommended by the \gls{ENFSI} in their guidelines for face image comparisons \cite{BestPracticeManual2018a}. 
Taken together, these documents provide a solid reference point for deriving requirements for explainable systems, as well as expectations for the explanations produced by such systems. They motivate both what should be described (morphology) and what should be avoided (unstable cues).

\subsection{Explainability of vision models}
On the machine learning side, early work on explainability in vision tasks has largely focused on post-hoc saliency methods such as Grad-CAM \cite{selvarajuGradCAMVisualExplanations2017}.
These approaches have also been applied to face recognition \cite{willifordExplainableFaceRecognition2020,linXCosExplainableCosine2021} with the goal of highlighting facial regions that most influence the predictions of automated \glspl{FRS}.

However, saliency maps have limitations highlighted in prior work that may strongly restrict their applicability in forensic settings.
In particular, while a saliency map can indicate \emph{where} a model is looking, it does not necessarily inform \emph{what} it is looking at.
For example, if the eye region is highlighted in a saliency map for face recognition, is it because pupil size is informative, because periocular skin texture is informative, or because glasses are (incorrectly) treated as informative?
Furthermore, the ambiguous nature of these heatmaps can inadvertently trigger confirmation bias, leading human examiners to interpret visualizations as supporting prior beliefs rather than objectively assessing whether the model relied on valid discriminative features \cite{adebayoSanityChecksSaliency2018a,alqaraawiEvaluatingSaliencyMap2020}.

Further efforts aim to mitigate these issues through the development of Concept Bottleneck Models \cite{kohConceptBottleneckModels2020}.
For image classification tasks, these models constrain the internal representation by first mapping images to a space of predefined, human-understandable concepts, and then performing classification starting only from that controlled space.
Unfortunately, such approaches require concept-labeled data at a scale and granularity that is not yet available for face recognition, especially given the level of fine-grained detail needed to satisfy forensic expectations.

\subsection{Vision-Language Models in Biometrics}
More recently, the emergence of \glspl{VLM} has opened a new line of research on explainable systems.
The rationale is that such models, in addition to performing vision tasks, can produce text-form rationales that may better match forensic expectations than saliency maps.
The biometrics community has started exploring the use of \glspl{VLM} for biometrics and face understanding.
Seminal works \cite{deandres-tameHowGoodChatGPT2024,hassanpourChatgptBiometricsAssessment2024} provide early and systematic studies of the multimodal variant of ChatGPT for face biometrics, covering face verification, soft biometrics, and general face understanding, together with an initial discussion of explainability aspects.
This has led to the development of multiple benchmark works \cite{otroshishahrezaFoundationModelsBiometrics2025,narayanFaceXBenchEvaluatingMultimodal2025,sonyBenchmarkingFoundationModels2025} that more broadly evaluate the capability of \glspl{VLM}, including open-weight models, to perform face verification and face understanding in zero-shot and few-shot settings.
Together, these benchmarks provide valuable evidence that inference-time prompting can yield non-trivial biometric performance without model fine-tuning.
While \glspl{VLM} typically do not match the performance of dedicated task-specific networks on these tasks, these works emphasize the added value of text-based interaction and text generation as a non-negligible benefit.
Efforts have also appeared to further improve \glspl{VLM}' performance through fine-tuning, such as FaceLLM \cite{shahrezaFaceLLMMultimodalLarge2025} for face understanding tasks.

A recurring theme across the above works is that multimodal foundation models can produce natural-language outputs, and can therefore be prompted to provide text-based rationales alongside biometric decisions.
This property is particularly appealing for forensic applications, where practitioners typically expect auditable, human-readable reporting rather than purely numerical scores.
However, existing benchmarks and evaluations primarily emphasize task performance (e.g., verification accuracy or task success rates) and typically discuss explanation quality only at a qualitative level.
As a result, although the literature increasingly demonstrates that \glspl{VLM} can both \emph{decide} and \emph{describe}, it remains unclear under which conditions the produced textual rationales are valid, stable, and aligned with forensic expectations.

In that regard, faithfulness metrics for \glspl{VLM} have been proposed in prior work \cite{jingFaithScoreFinegrainedEvaluations2024}, but their reliance on an LLM-as-a-judge to convert free text into scorable items introduces an additional failure mode via judge-model errors, which is problematic.
Instead, we prefer more constrained text production from the benchmarked \gls{VLM}, enabling scoring through more predictable heuristics.
Relevance metrics are typically task-specific, and there is therefore a need to establish such a metric for face recognition applications.

The aim of this work is thus to advance the application of \glspl{VLM} to face recognition by proposing a methodology that properly quantifies the quality of generated explanations, specifically with the goal of providing content useful to forensic experts.

\section{Methodology}
\begin{figure}[htbp]
\centering
{\small
\begin{tabularx}{\linewidth}{|X|}
\hline
\textbf{System Prompt} \\
You are a face identity verification expert. \\
\hline
\textbf{User Prompt} \\
Analyze the two provided face images (A and B).
Compare them feature-by-feature to decide whether they show the same person or different people.
Output a single JSON object with exactly these keys:
\begin{itemize}
    \setlength\itemsep{0em}
    \item \texttt{feature\_comparisons}: list of facial feature comparisons; for each feature include:
    \begin{itemize}
        \item \texttt{feature}: the facial feature name in snakecase format
        \item \texttt{description\_a}: description of the feature in image A.
        \item \texttt{description\_b}: description of the feature in image B.
        \item \texttt{comparison}: concise comparison of this feature across A and B.
    \end{itemize}
    \item \texttt{explanation}: a brief global justification summarizing how the local comparisons lead to the final decision.
    \item \texttt{similarity\_score}: a number between 0.0 (definitely different people) and 1.0 (definitely the same person).
\end{itemize}
Do not nominally mention the names of the individuals in the images; focus only on facial characteristics. Respond strictly in JSON format with the specified keys. \\
\hline
\end{tabularx}
}
\caption{The prompts used for the structured setting}
\label{fig:prompts}
\end{figure}

\subsection{Datasets, models, prompting}
We study face verification in a VLM-based setting: given a pair of face images $(I_a, I_b)$, a VLM is prompted to output a global similarity score $s \in [0,1]$. In addition to verification performance, we evaluate the quality of the generated explanations along two axes motivated by forensic reporting needs: \textbf{relevance} and \textbf{faithfulness}.

We evaluate on four datasets: LFW \cite{huangLabeledFacesWild2008}, which is chosen as a commonly used baseline, ARFace \cite{martinezARFaceDatabase1998} and Soteria \cite{ramolyNovelResponsibleDataset2024}, which contain occlusion-specific annotations enabling faithfulness measurements, and CelebA \cite{liuDeepLearningFace2015} to experiment at larger scale and in more diverse conditions.
For LFW, we use the standard 6000-pair protocol (3000 genuine, 3000 impostor).
For ARFace, Soteria, and CelebA, we construct LFW-like protocols with (i) balanced genuine/impostor pairs, and (ii) 10-fold identity-disjoint splits to avoid identity leakage between folds.
We detect and crop faces using the InsightFace library \cite{dengRetinaFaceSingleShotMultiLevel2020}. When face detection fails on at least one image from a pair, the entire pair is treated as a failure-to-acquire (FTA) and excluded from metric computation.

For benchmarking in a zero-shot setting, we select Gemma3~\cite{teamGemma3Technical2025}, Qwen2.5-VL~\cite{baiQwen25VLTechnicalReport2025}, and InternVL3~\cite{zhuInternVL3ExploringAdvanced2025} as three recent, widely-adopted open-weight VLM families spanning 1B--78B parameters, enabling a scaling analysis across distinct training recipes and vision backbones. We run them with deterministic decoding (temperature $0$) and fixed seed to ensure reproducibility.
We compare two output modes:
\begin{itemize}
\item{\textbf{Unstructured explanations}:
The model is prompted to return (i) a float similarity score $s \in [0,1]$ and (ii) a free-form textual justification. This is the type of setting observed in prior works.}
\item{\textbf{Structured explanations}: 
We constrain outputs to a fixed JSON schema enforced at token sampling time using constrained decoding. The prompt is presented in Figure \ref{fig:prompts}, and a corresponding JSON schema is applied to the model's output to ensure compliance. Importantly, note that the number of feature comparisons, and the choice of compared features, is left completely free to the model. Small models occasionally fail to satisfy the schema; such cases are counted as FTAs.
Figure \ref{fig:project schematic} presents an example of the output produced by the models in the structured setting.
}
\end{itemize}
\subsection{Evaluation metrics}

\subsubsection{Verification performance}
Using the similarity score $s$, we report (i) \gls{EER}, as well as (ii) accuracy which is commonly used in prior VLM-for-face-verification literature. Given that VLMs are not yet at the level of operational performance, we do not report false accept / false reject rates at specific thresholds, which would not be very informative in the current case. 
For accuracy, we follow a 10-fold protocol: on each fold, we select a threshold on the 9 remaining folds (e.g., maximizing accuracy) and evaluate on the held-out fold, then average across folds.
We additionally report the FTA rate, counting failures due to face detection or invalid structured outputs.

\subsubsection{Explanation quality metrics}
We propose a quantification of two additional criteria:
\paragraph{Relevance (identity-stable vs transient cues)}
Relevance measures whether explanations emphasize \emph{identity-stable} facial evidence rather than transient or forensically unstable cues (e.g., glasses, facial expression, illumination).
We define a lexicon $\mathcal{L}$ of unstable cues and measure how often they are invoked in the \texttt{feature} field of produced explanations.

We report the \textbf{unstable-feature rate}
\[
U = \frac{\#\{\text{\texttt{feature} entries mentioning a cue in } \mathcal{L}\}}{\#\{\text{all \texttt{feature} entries}\}},
\]
and the \textbf{relevance score} as its complement $R = 1 - U$.
\begin{table}[htbp]
    \centering
    \small
    \caption{Lexicons used for Relevance and Faithfulness metrics. Counts in parentheses indicate the total number of keywords per category. Note on occlusion acknowledgments: if a listed feature appears in the trigger lexicon but its description simply acknowledges occlusion (by containing these terms), we do not count it as a hallucination. The full lexicon is presented as supplementary material.}
    \label{tab:lexicons_compact}
    \begin{tabularx}{\linewidth}{lX}
    \toprule
    \textbf{Category} & \textbf{Example keywords} \\
    \midrule
    \multicolumn{2}{c}{\textit{Relevance metric: unstable features}} \\
    \midrule
    Accessories (27) & glasses, mask, hat, scarf, jewelry, ... \\
    Attire (17) & shirt, jacket, collar, sleeve, outfit, ... \\
    Environment (16) & background, lighting, shadow, blur, image, ... \\
    Expression (15) & expression, smile, gaze, frown, gesture, ... \\
    Grooming (16) & makeup, hairstyle, beard, mustache, blemish, ... \\
    \midrule
    \multicolumn{2}{c}{\textit{Faithfulness metric: trigger words for facial regions}} \\
    \midrule
    Eyes (8) & eye, iris, pupil, eyelid, periocular, ... \\
    Nose (5) & nose, nostril, nasal, bridge, ... \\
    Mouth (7) & mouth, lips, teeth, smile, grin, ... \\
    \midrule
    \multicolumn{2}{c}{\textit{Faithfulness metric: occlusion acknowledgments}} \\
    \midrule
    Indicators (19) & occluded, covered, hidden, masked, obscured, not visible, hard to tell, partially visible, ... \\
    \bottomrule
    \end{tabularx}
\end{table}

\paragraph{Faithfulness (no hallucinations beyond visible evidence)}
Faithfulness measures whether explanations respect what is actually visible in the images and avoid hallucinating details.
For practical reasons, we propose a specialized version of faithfulness focused only on occlusion considerations, by penalizing mentions of facial parts known to be occluded. This can be the eyes or mouth area in the case of the ARFace dataset, and the nose or mouth area in the case of the Soteria dataset.
Concretely, we again define a lexicon of the facial features which are occluded for each type of considered occlusion.  Together with available occlusion labels from the datasets, we can flag an explanation item as \emph{hallucinated} if the \texttt{feature} field refers to a facial part occluded in either $I_a$ or $I_b$.
We report the \textbf{hallucination rate}
\[
H = \frac{\#\{\text{\texttt{feature} entries referring to occluded parts}\}}{\#\{\text{all \texttt{feature} entries}\}},
\]
and faithfulness as $F = 1 - H$.

For both relevance and faithfulness, we report a macro-average across all comparisons: per-comparison rates are computed over the variable-length feature list, then averaged across comparisons.

Table \ref{tab:lexicons_compact} presents the lexicons that are used in practice for our experiments.

All experiments are run in PyTorch \cite{paszkePyTorchImperativeStyle2019a} with vLLM \cite{kwonEfficientMemoryManagement2023} for inference and constrained decoding. We will release prompts, JSON schemas, and evaluation code\footnote{URL will be provided upon acceptance.}
 to support reproducibility and to serve as an evaluation harness for future fine-tuning of VLMs for face verification with auditable explanations.

\section{Results}
\begin{table*}[htbp]
\caption{\gls{EER}, Accuracy (ACC) and \gls{FTA}, in \% in the structured output setting. The best performing VLM (in EER, respectively accuracy) is \textbf{bolded}.}
\label{tab:eer}
\resizebox{\linewidth}{!}{%
\begin{tabular}{@{}llrrrrrrrrrrrr@{}}
    \toprule
                                   & Dataset & \multicolumn{3}{c}{\cellcolor{gray!15}LFW} & \multicolumn{3}{c}{\cellcolor{gray!35}ARFace} & \multicolumn{3}{c}{\cellcolor{gray!15}Soteria} & \multicolumn{3}{c}{\cellcolor{gray!35}CelebA}                                                          \\
                                   & Metric  & EER                                        & ACC                                           & FTA                                            & EER                                           & ACC   & FTA  & EER  & ACC  & FTA  & EER  & ACC  & FTA  \\
    \midrule
    ArcFace                        & -       & 0.3                                        & 99.9                                          & 0.0                                            & 0.1                                           & 100.0 & 0.0  & 0.3  & 99.7 & 0.1  & 3.0  & 98.1 & 0.2  \\
    \cline{1-14}
    \multirow[t]{3}{*}{Gemma3}     & 4B      & 9.5                                        & 90.4                                          & 0.1                                            & 11.3                                          & 88.9  & 0.1  & 9.0  & 91.5 & 0.2  & 13.5 & 86.5 & 0.2  \\
                                   & 12B     & 7.6                                        & 92.6                                          & 0.0                                            & 17.1                                          & 85.1  & 0.0  & 10.1 & 90.7 & 0.1  & 12.1 & 88.3 & 0.2  \\
                                   & 27B     & \bfseries 5.2                                        & 94.7                                          & 0.0                                            & \bfseries7.1                                           & \bfseries93.0  & 0.0  & \bfseries4.6  & \bfseries95.5 & 0.1  & \bfseries10.4 &\bfseries 89.6 & 0.2  \\
    \cline{1-14}
    \multirow[t]{4}{*}{Qwen2.5-VL} & 3B      & 44.3                                       & 55.5                                          & 2.6                                            & 39.8                                          & 60.3  & 1.8  & 38.3 & 63.0 & 2.2  & 44.2 & 55.7 & 2.6  \\
                                   & 7B      & 25.1                                       & 75.0                                          & 0.4                                            & 28.4                                          & 71.5  & 0.3  & 26.1 & 75.0 & 0.4  & 28.0 & 71.9 & 0.4  \\
                                   & 32B     & 8.0                                        & 92.1                                          & 0.0                                            & 14.4                                          & 86.2  & 0.0  & 10.5 & 90.1 & 0.1  & 11.1 & 89.4 & 0.2  \\
                                   & 72B     & \bfseries 5.2                                        & \bfseries 94.9                                          & 0.0                                            & 12.6                                          & 87.6  & 0.0  & 7.6  & 93.0 & 0.1  & 11.6 & 88.6 & 0.2  \\
    \cline{1-14}
    \multirow[t]{5}{*}{InternVL3}  & 1B      & 48.9                                       & 51.6                                          & 28.7                                           & 49.0                                          & 56.0  & 26.6 & 48.9 & 52.5 & 26.8 & 48.7 & 51.6 & 29.2 \\
                                   & 8B      & 9.8                                        & 90.3                                          & 0.0                                            & 12.3                                          & 87.8  & 0.0  & 9.6  & 90.8 & 0.1  & 17.6 & 83.0 & 0.2  \\
                                   & 14B     & 6.2                                        & 93.7                                          & 0.0                                            & 9.8                                           & 90.5  & 0.0  & 7.0  & 93.6 & 0.1  & 13.3 & 86.9 & 0.2  \\
                                   & 38B     & 5.3                                        & 94.5                                          & 0.0                                            & 10.1                                          & 90.7  & 0.0  & 5.6  & 94.9 & 0.1  & 13.2 & 86.9 & 0.2  \\
                                   & 78B     & 5.8                                        & 94.2                                          & 0.0                                            & 11.8                                          & 88.9  & 0.0  & 6.3  & 94.5 & 0.1  & 12.7 & 87.4 & 0.2  \\
    \bottomrule
\end{tabular}

}
\end{table*}

\begin{table*}[htbp]
\caption{\textbf{Difference} in \gls{EER}, Accuracy and \gls{FTA}, in percentage points, in the structured output setting compared to the unstructured output setting. A positive value means that the corresponding metric is \emph{higher} in the structured output setting than in the unstructured one.}
\label{tab:eer_delta}
\resizebox{\linewidth}{!}{%
\begin{tabular}{@{}llrrrrrrrrrrrr@{}}
\toprule
 & Dataset & \multicolumn{3}{c}{\cellcolor{gray!15}LFW} & \multicolumn{3}{c}{\cellcolor{gray!35}ARFace} & \multicolumn{3}{c}{\cellcolor{gray!15}Soteria} & \multicolumn{3}{c}{\cellcolor{gray!35}CelebA} \\
 & Metric & EER  & ACC  & FTA  & EER  & ACC  & FTA  & EER  & ACC  & FTA  & EER  & ACC  & FTA  \\
\midrule
\multirow[t]{3}{*}{Gemma3} & 4B & +2.6 & -2.8 & +0.1 & +1.5 & -1.4 & +0.1 & +2.9 & -2.8 & +0.0 & +1.4 & -1.4 & +0.1 \\
 & 12B & -2.4 & +0.5 & -0.0 & +0.2 & +5.6 & -0.0 & -4.0 & +2.7 & +0.0 & -1.5 & +0.4 & -0.0 \\
 & 27B & +1.1 & -1.2 & -0.0 & +0.6 & +0.1 & -0.0 & +0.2 & -0.1 & +0.0 & +1.2 & -1.2 & -0.0 \\
\cline{1-14}
\multirow[t]{4}{*}{Qwen2.5-VL} & 3B & +14.1 & -16.2 & +2.5 & +8.5 & -10.7 & +1.8 & +8.0 & -7.5 & +2.0 & +12.9 & -13.5 & +2.4 \\
 & 7B & +19.0 & -18.9 & +0.3 & +17.1 & -17.8 & +0.2 & +15.2 & -15.2 & +0.2 & +14.3 & -15.9 & +0.1 \\
 & 32B & +2.9 & -2.6 & +0.0 & +3.2 & -3.0 & -0.0 & +2.5 & -3.2 & +0.0 & +0.6 & -0.9 & +0.0 \\
 & 72B & +0.4 & -0.7 & -0.0 & +4.2 & -3.7 & -0.0 & +3.5 & -2.8 & +0.0 & +0.8 & -0.5 & +0.0 \\
\cline{1-14}
\multirow[t]{5}{*}{InternVL3} & 1B & +1.7 & -2.2 & +23.0 & +1.3 & -0.4 & +22.4 & +2.4 & -1.5 & +21.5 & +1.4 & -1.3 & +24.2 \\
 & 8B & +3.6 & -3.5 & +0.0 & +3.1 & -3.7 & +0.0 & +2.1 & -2.3 & -0.0 & +4.6 & -3.9 & -0.0 \\
 & 14B & -1.6 & +1.3 & -0.0 & +3.6 & -3.4 & -0.0 & +1.8 & -1.1 & +0.0 & -2.1 & +2.2 & -0.0 \\
 & 38B & +0.9 & -1.4 & -0.0 & +2.4 & -2.4 & +0.0 & +1.3 & -0.9 & -0.0 & +0.7 & -0.7 & +0.0 \\
 & 78B & +1.9 & -1.6 & -0.0 & +2.6 & -2.7 & -0.0 & +2.6 & -2.0 & -0.0 & +0.6 & -1.5 & -0.0 \\
\bottomrule
\end{tabular}

}
\end{table*}
\subsection{Verification performance in the structured setting}
\begin{figure*}[htbp]

\begin{subfigure}{\textwidth}
\includegraphics[width=\linewidth]{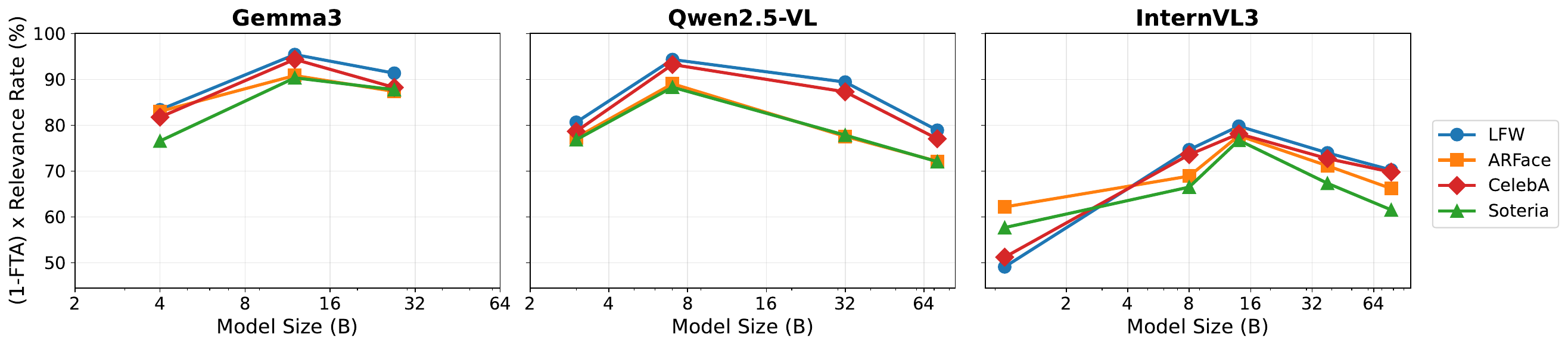}
\vspace{-0.4cm}

\label{fig:relevance_scaling}
\end{subfigure}
\begin{subfigure}{\textwidth}
\includegraphics[width=\linewidth]{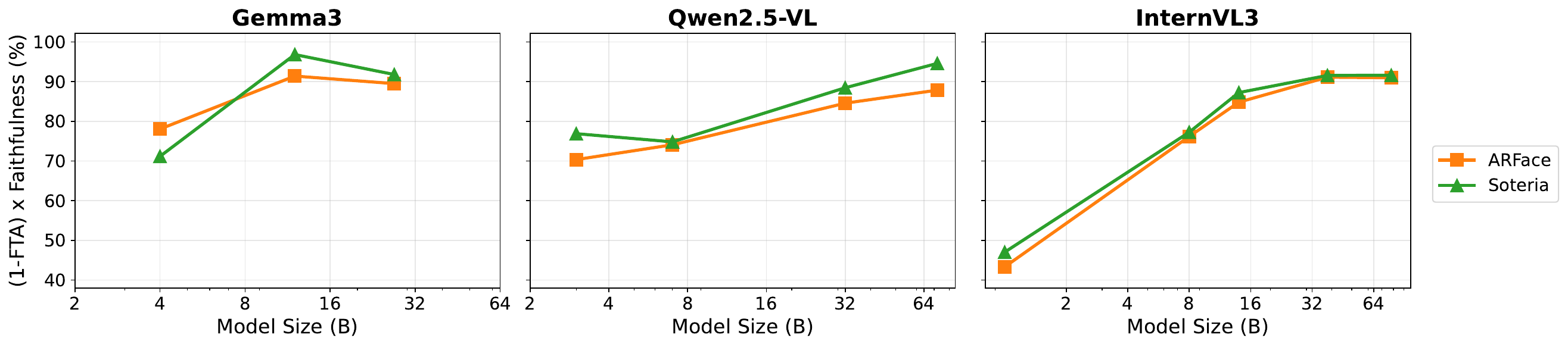}
\label{fig:faithfulness_scaling}
\end{subfigure}
\caption{Scaling trends for relevance and faithfulness metrics. We report $(1-\mathrm{FTA}) \times$ score for relevance and faithfulness, reflecting end-to-end usable explanation quality by accounting for cases where no valid structured explanation is produced (FTA).}
\label{fig:scaling}
\end{figure*}
Table~\ref{tab:eer} reports EER, accuracy, and failure-to-acquire (FTA) rates when enforcing structured JSON output. As a reference point, we also report the performance of a traditional state-of-the-art face recognition model. We use the Buffalo-L model pack from the  InsightFace library, which is a ResNet50 trained on WebFace12M \cite{zhuWebFace260MBenchmarkMillionScale2023} with the ArcFace loss \cite{dengArcFaceAdditiveAngular2022}, and the RetinaFace face detector. Overall, VLM-based verification remains substantially behind dedicated face recognition systems, and performance varies across both model families and scales. In particular, Gemma3-27B achieves the lowest EERs across most datasets (e.g., 5.2\% on LFW and 7.1\% on ARFace), while Qwen2.5-VL and InternVL3 exhibit stronger dependence on scale, with small sizes yielding very high error rates (e.g., Qwen2.5-VL-3B and InternVL3-1B near chance-level EERs). Table~\ref{tab:eer} also highlights that structured decoding introduces additional failure modes: smaller models can fail to comply with the required JSON schema, resulting in elevated FTA (e.g., InternVL3-1B with $\approx$27--29\% FTA across datasets). These failures are excluded from metric computation but are reported explicitly as an operational cost of structured explanations.

Across families, scaling models up generally improves verification performance, but not systematically. For instance, Gemma3-12B underperforms Gemma3-4B on ARFace and Soteria, and InternVL3-78B is slightly worse than InternVL3-38B on LFW, ARFace and Soteria. 
 Deviations in performance across families suggest that, beyond parameter count, family-specific training recipes and vision backbones can strongly impact face verification behavior.

\subsection{Cost of structured output constraints}
Table~\ref{tab:eer_delta} reports the difference in EER, accuracy, and FTA between structured and unstructured output modes. In most settings, enforcing structured explanations increases EER, indicating a recognition cost associated with constrained generation. This cost is particularly severe for smaller models, where the dominant issue is often the FTA increase caused by schema non-compliance rather than a pure degradation of similarity scoring. As models scale up, the gap between structured and unstructured performance decreases, suggesting that larger VLMs better tolerate output constraints and that the trade-off between recognition performance and auditable explanations becomes negligible at scale.

Interestingly, a few configurations yield negative $\Delta$EER (i.e., structured output slightly lowers EER), such as Gemma3-12B on LFW and Soteria. While this effect is not consistent across datasets, it suggests that structured decomposition can sometimes act as a regularizer by forcing models to perform a more explicit feature-by-feature comparison before producing a global score.

\subsection{Scaling trends: accuracy vs.\ explanation quality}
From Table \ref{tab:eer} we also observe that the EER generally decreases with scale, with consistent trends across datasets and families, albeit with the non-monotonic exceptions discussed above.
Figure \ref{fig:scaling} presents similar scaling trends for the relevance and faithfulness metrics. It demonstrates that the proposed metrics reveal differences between models that are not visible from EER alone. For example, InternVL3-38B and Qwen2.5-VL-72B show comparable verification performance, yet InternVL3-38B exhibits substantially lower relevance, indicating a stronger tendency to rely on forensically unstable cues in its explanations. In other words, similar match performance can correspond to very different explanation behaviors, motivating explanation quality as a complementary evaluation axis. We observe substantial remaining shortcomings: some models such as InternVL3 mention forensically unstable cues in up to ~20\% of feature entries, and refer to occluded regions in up to ~10\% of entries on occlusion datasets.
Relevance scaling exhibits a notable pattern: the best relevance is typically achieved by intermediate-sized models (Gemma3-12B, Qwen2.5-VL-7B, InternVL3-14B in our benchmark), and relevance often decreases for the largest sizes. This suggests that scaling can increase the frequency with which models invoke transient contextual cues (e.g., accessories, expression, image quality) when generating explanations. In contrast, faithfulness tends to improve more consistently with scale on occlusion-focused datasets, indicating that larger models are less prone to describing facial parts that are not visible under occlusions, although we again observe exceptions such as Gemma3-27B performing worse than Gemma3-12B.

\subsection{Automatic audits: unstable cues and hallucinations}
Table~\ref{tab:relevance_examples} provides examples of the most frequently observed unstable cues. It illustrates the practical value of a structured approach that enables automated auditing. Reliance on hair and facial hair is common; while these may remain usable cues in some settings, they are more easily altered than stable facial morphology and can be problematic depending on the context. More clearly invalid cues (e.g., glasses/eyewear, expression, lighting, accessories, background, clothing) also occur frequently, despite being transient or irrelevant to identity. We additionally observe multiple cases where models produce nearly identical descriptions for images A and B, which may reflect a sequential generation bias where the first description influences the second. Table~\ref{tab:relevance_examples} further highlights occasional degenerate outputs (e.g., a description field containing an unrelated proper name), suggesting that additional validation checks may be beneficial beyond schema compliance.

\begin{figure}[htbp]
    \centering
    \begin{subfigure}[t]{0.32\columnwidth}
        \centering
        \includegraphics[width=\textwidth]{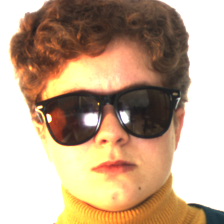}
        \caption{Dark-colored eyes.}
        \label{fig:example_8783}
    \end{subfigure}
    \hfill
    \begin{subfigure}[t]{0.32\columnwidth}
        \centering
        \includegraphics[width=\textwidth]{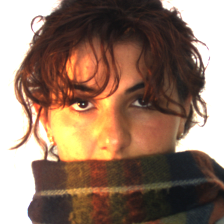}
        \caption{Thicker lips}
        \label{fig:example_3167}
    \end{subfigure}
    \hfill
    \begin{subfigure}[t]{0.32\columnwidth}
        \centering
        \includegraphics[width=\textwidth]{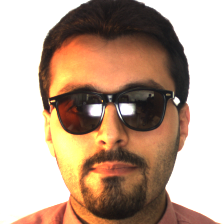}
        \caption{Darker eyes, slightly wider-set.}
        \label{fig:example_288}
    \end{subfigure}
    \vspace{1em}
    \begin{subfigure}[t]{0.32\columnwidth}
        \centering
        \includegraphics[width=\textwidth]{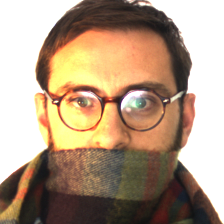}
        \caption{Slightly wider mouth with thinner lips.}
        \label{fig:example_9774}
    \end{subfigure}
    \hfill
    \begin{subfigure}[t]{0.32\columnwidth}
        \centering
        \includegraphics[width=\textwidth]{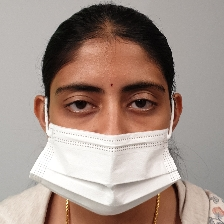}
        \caption{Image B also shows a straight, medium-sized nose.}
        \label{fig:example_6102}
    \end{subfigure}
    \hfill
    \begin{subfigure}[t]{0.32\columnwidth}
        \centering
        \includegraphics[width=\textwidth]{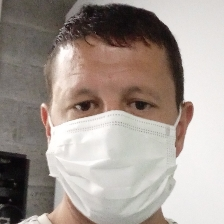}
        \caption{Medium-sized nose, but appears narrower with a less broad bridge.}
        \label{fig:example_11021}
    \end{subfigure}

    \caption{Examples of hallucinations in facial descriptions. The caption is the content of the \texttt{description} field produced by the VLM for that specific image.}
    \label{fig:faithfulness_examples}
\end{figure}

Figure~\ref{fig:faithfulness_examples} illustrates faithfulness failures on occlusion datasets: models sometimes describe attributes that are not visible due to sunglasses, scarves, or masks. Some errors can be interpreted as plausible confusions (e.g., sunglasses prompting ``dark eyes''), but others are unsupported fabrications (e.g., detailed nose-shape claims when the lower face is masked). A potential contributing factor is again the sequential nature of generation: in our protocols, the first image is often not occluded while the second may be occluded, which can encourage the model to introduce a feature based on the first image and then continue describing it for the second even when it is not observable.

\begin{table}[htbp]
    \centering
    \caption{Examples of unstable features in face descriptions. We showcase a mixture of examples from all models and all datasets.}
    \label{tab:relevance_examples}
    {\scriptsize
    \begin{tabular}{@{}p{0.13\columnwidth}p{0.37\columnwidth}p{0.37\columnwidth}@{}}
        \toprule
        \textbf{Feature} & \textbf{Description A} & \textbf{Description B} \\
        \midrule
        hair & short, grey hair & no detectable hair \\
        hairstyle & Long, straight hair framing the face. & Short, dark hair styled back. \\
        glasses & no glasses & black expansive glasses with reflective lenses \\
        beard & Full, graying beard and mustache. & Full, graying beard and mustache. \\
        expression & Neutral expression. & Slightly intense expression. \\
        eyewear & thick rim glasses with noticeable reflection & thin rim glasses with noticeable reflection \\
        smile & smile & Macie Taylor \\
        earring & Present on both ears & Missing \\
        background & Indoor setting with a cardboard box. & Outdoor setting with trees and grass. \\
        cloth & white shirt with black horizontal stripes & plaid pattern shirt \\
        jewelry & glasses worn & no visible glasses \\
        lighting & Fairly even lighting, with some glare. & Bright sunlight. \\
        shirt & light blue striped & light blue striped \\
        \bottomrule
    \end{tabular}
    }
\end{table}

\section{Discussion}
\subsection{Findings and implications}
This work introduces a benchmarking methodology for VLM-based face verification that evaluates not only recognition accuracy but also two forensic-motivated explanation quality criteria: relevance and faithfulness. Across three open-weight VLM families and multiple scales, we observe (i) a clear gap between VLM-based verification and dedicated face recognition systems, (ii) a consistent recognition cost associated with structured output constraints, which diminishes with model scale, and (iii) significant variations in explanation quality that are not predictable from EER alone. In particular, models with similar EER can differ sharply in their reliance on transient cues (relevance) and their tendency to hallucinate occluded facial parts (faithfulness).
From a practitioner perspective, these results suggest that selecting a VLM for explanatory face verification should not be based solely on match performance. Larger models generally reduce both EER and schema-related FTA, but scaling does not reliably improve relevance, and intermediate-sized models can produce more forensically appropriate explanations even when their EER is slightly worse. Model choice therefore depends on the application goal: larger models offer better robustness when verification accuracy is the primary objective, while intermediate sizes may be preferable when minimizing reliance on transient cues is critical.
The qualitative audits also highlight concrete failure modes that matter for forensic reporting, including frequent reliance on accessories, expression, illumination, or image quality, as well as descriptions of occluded facial regions. These provide clear guidance for further model refinement. The proposed framework additionally enables a complementary use case: when fine-tuning VLMs to improve face recognition performance, explanation quality can be monitored to detect unintended regressions.

\subsection{Limitations}
Our relevance metric relies on lexicon-based detection of unstable cues and therefore captures a conservative subset of relevance violations; further effort would be necessary to ensure completeness of the lexicon.  Faithfulness is restricted to focusing on occlusion annotations (ARFace and Soteria); it does not capture hallucinations unrelated to occlusion (e.g., invented scars) or errors on non-occluded datasets. The restriction to deterministic decoding prevents measuring the variability of the evaluated metrics, and we also do not quantify the effect of alternative prompting strategies or non-zero temperatures. Our verification protocol uses LFW, ARFace, Soteria, and CelebA. While ARFace and Soteria provide controlled occlusion conditions relevant to our faithfulness analysis, LFW and CelebA contain publicly available web imagery that may overlap with VLM pretraining data, so reported results may partially conflate generalization with memorization. Extending the evaluation to harder verification benchmarks (e.g., CFP-FP, AgeDB-30, CA-LFW, CP-LFW, IJB-B/C) would provide a more complete picture of VLM robustness and is left to future work. Finally, while this paper focuses on relevance and faithfulness, additional criteria are likely to be necessary for gaining a complete picture. In particular,  we believe the evaluation of \textbf{correctness} (correspondence between VLM fine-grained facial attributes claims on visible, stable features, and the visual ground truth) remains an important direction, but is more challenging due to the need for associated labels of a level of fineness uncommon in available datasets.

\subsection{Future work}
Several extensions follow from this benchmark. Stricter prompting and output constraints could be explored, e.g., restricting admissible features to a forensically-aligned lexicon, or extending the schema with explicit ``visibility'' or ``confidence'' fields enabling models to abstain when evidence is occluded. The schema could specifically be designed to follow FISWG morphological analysis guidelines, mapping explanations directly to forensic reporting structure. Fine-tuning models with objectives that explicitly target relevance and faithfulness (rather than verification accuracy alone) is another natural direction, with the automated metrics proposed here serving to validate explanation quality during training. Finally, user studies with forensic practitioners would be valuable to validate whether automated improvements correspond to perceived usefulness in casework reporting.

\section{Conclusion}
We presented a benchmarking methodology for VLM-based face verification that evaluates not only recognition accuracy, but also two forensic-motivated explanation quality criteria: relevance and faithfulness. By constraining model outputs to a structured format, our approach enables automated auditing of explanations at scale. Across three open-weight VLM families and multiple sizes, we observed a clear gap with dedicated FRSs, a measurable cost of structured constraints diminishing with scale, and substantial differences in explanation behavior between models with similar EER. We will release our benchmarking framework to enable future work on improving VLMs while preserving auditable explanations.

\section*{Acknowledgments}
This work was supported by the Center of Identification Technology Research (CITeR) and the Idiap Research Institute.

{\small
\bibliographystyle{ieee}
\bibliography{vlmfr}
}

\clearpage
\appendix

\section{Lexicons}
Complete lexicons used in the presented experiments for flagging hallucinations or exploitation of irrelevant features.

\begin{table}[htbp]
    \centering
    \small
    \caption{Lexicons used for Relevance and Faithfulness metrics. Note on the "occlusion acknowledgments" list: if a listed feature is part of the trigger lexicon, but the description of that feature simply acknowledges occlusion (by containing these terms), we do not count it as a hallucination. }
    \label{tab:lexicons}
    \begin{tabularx}{\linewidth}{lX}
    \toprule
    \textbf{Category} & \textbf{Keywords} \\
    \midrule
    \multicolumn{2}{c}{\textit{Relevance metric: unstable features }} \\
    \midrule
    Accessories & glasses, spectacles, eyewear, goggles, sunglasses, mask, respirator, shield, visor, hat, cap, helmet, beanie, hood, headwear, bandana, headband, scarf, tie, jewelry, earring, piercing, stud, hoop, necklace, clip, bindi \\
    Attire & cloth, shirt, jacket, coat, attire, garment, outfit, dress, suit, uniform, collar, sleeve, zipper, neckline, shoulder, chest, glove \\
    Environment & background, foreground, backdrop, scene, context, lighting, shadow, blur, focus, quality, noise, pixel, image, photo, picture, crop \\
    Expression & expression, emotion, mood, feeling, gaze, look, smile, smiling, laugh, frown, grimace, mouth open, teeth visible, wink, gesture \\
    Grooming & makeup, cosmetic, mascara, liner, lipstick, hairstyle, haircut, hairdo, dye, beard, mustache, shaved, trim, acne, blemish, pimple \\
    \midrule
    \multicolumn{2}{c}{\textit{Faithfulness metric: trigger word for facial regions}} \\
    \midrule
    Eyes & eye, eyes, iris, pupil, eyelid, eyelash, eyelashes, periocular \\
    Nose & nose, nostril, nostrils, nasal, bridge \\
    Mouth & mouth, lips, lip, teeth, smile, smiling, grin \\
    \midrule
    \multicolumn{2}{c}{\textit{Faithfulness metric: occlusion acknowledgments}} \\
    \midrule
    Indicators & occluded, covered, hidden, hiding, masked, masking, obscured, obscuring, not visible, not fully visible, not observable, cannot see, can't see, hard to tell, unclear, not clear, out of frame, blocked, partially visible \\
    \bottomrule
    \end{tabularx}
\end{table}

\clearpage
\pagenumbering{arabic}

\end{document}